\documentclass[10pt,twocolumn,letterpaper]{article}

\usepackage{cvpr}
\usepackage{times}
\usepackage{epsfig}
\usepackage{graphicx}
\usepackage{amsmath}
\usepackage{amssymb}
\usepackage[table,xcdraw]{xcolor}
\usepackage{adjustbox}
\usepackage{multirow}
\usepackage{floatrow}
\usepackage{booktabs}
\usepackage{float}

\usepackage[breaklinks=true,bookmarks=false]{hyperref}
\DeclareMathOperator{\EX}{\mathbb{E}}

\cvprfinalcopy 

\def\cvprPaperID{12} 

\usepackage{xcolor} 
\usepackage{soul} 
\usepackage{xspace}
\usepackage{amsmath}

\usepackage{amssymb}
\usepackage{pifont}

\newcommand{\cmark}{\ding{51}}%
\newcommand{\xmark}{\ding{55}}%

\usepackage{amsmath}

\definecolor{Green}{HTML}{AFDAAF}

\begin{document}

\title{PoliTO-IIT Submission to the EPIC-KITCHENS-100 Unsupervised Domain Adaptation Challenge for Action Recognition}

\author{Chiara Plizzari{\thanks{The authors equally contributed to this work. This paper is partially supported by the ERC project RoboExNovo. We also acknowledge
that the research activity herein was carried out using the IIT
HPC infrastructure.}} $^{, 1}$ \quad
Mirco Planamente\footnotemark[1] $^{, 1,2}$ \quad 
Emanuele Alberti $^{1}$ \quad

Barbara Caputo\textsuperscript{1,2} \\

\and \textsuperscript{1} Politecnico di Torino\\
{\tt\small {name.surname}@polito.it}

\and \textsuperscript{2} Istituto Italiano di Tecnologia\\
{\tt\small {name.surname}@iit.it}
}

\maketitle

\begin{abstract}
In this report, we describe the technical details of our submission to the EPIC-Kitchens-100 Unsupervised Domain Adaptation (UDA) Challenge in Action Recognition. 
To tackle the domain-shift which exists under the UDA setting, we first exploited a 
recent Domain Generalization (DG) technique, called
Relative Norm Alignment (RNA). It consists in designing a model able to generalize well to any unseen domain, regardless of the possibility to access target data at training time.
Then, in a second phase, we extended the approach to work on unlabelled target data, allowing the model to adapt to the target distribution in an unsupervised fashion. For this purpose, we included in our framework existing UDA algorithms, such as Temporal Attentive Adversarial Adaptation Network (TA$^3$N), jointly with new multi-stream consistency losses, namely Temporal Hard Norm Alignment (T-HNA) and Min-Entropy Consistency (MEC).
Our submission (entry ‘plnet') is visible on the leaderboard and it achieved the 1st  position for \textit{‘verb’}, and the 3rd position for both \textit{‘noun’} and \textit{‘action’}.

\end{abstract}
\section{Introduction}

First person action recognition offers a wide range of opportunities which arise from the use of wearable devices. In fact, since it intrinsically comes with rich sound information, due to the strong hand-object interactions and the closeness of the sensors to the sound source, it encourages the use of auditory information. Moreover, the continuous movement of the camera, which moves around with the observer, strongly motivates the use of secondary modalities capturing the motion in the scene, such as optical flow. 

\begin{figure}[t]
    \centering
    \includegraphics[width=\columnwidth]{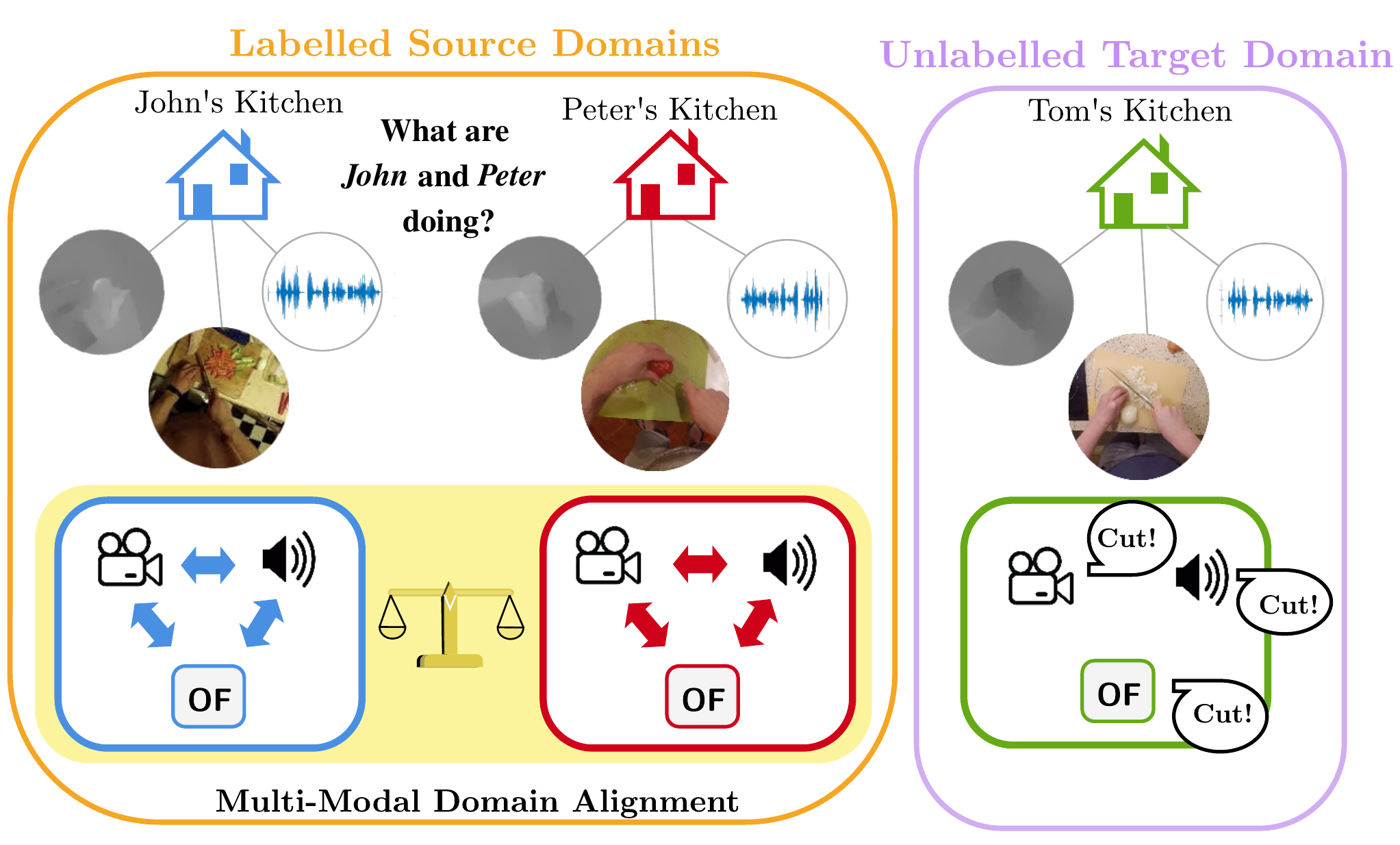}
    \caption{The correlation between the distinctive sound of an action and its corresponding visual information or motion is not always guaranteed across different domains. 
    Thus, effectively combining multi-modal information from \emph{multiple sources} is fundamental to increase the capability to recognize daily actions.}
    \label{fig:teaser}
\end{figure}

Our idea is that exploiting the intrinsic peculiarities of all these modalities is of crucial importance, especially in cross-domain scenarios. In fact, these modalities suffer from a domain shift which is not of the same nature. For instance, the optical flow modality, by focusing on the motion in the scene rather than on the appearance, is less sensitive to environmental changes, and thus potentially more robust than the visual modality when changing environment~\cite{munro2020multi} (Figure \ref{fig:teaser}). On the other side, the domain shift of auditory information is very different from the visual one (e.g., the sound of ‘cut' will differ from a plastic to a wooden cutting board). For all those reasons, the classifier should be able to measure and understand which modality is informative and should rely on in the final prediction, and which is not.

To this purpose, authors of \cite{planamente2021crossdomain} recently proposed a multi-modal framework, called Relative Norm Alignment network (RNA-Net), which aims to progressively align the feature norms of audio and visual (RGB) modalities among multiple sources in a Domain Generalization (DG) setting, where target data are not available during training. In that work, they bring to light that \emph{simply feeding all the source domains to the network without applying any adaptive techniques leads to sub-optimal performance. Indeed, a multi-source domain alignment allows the network to promote domain-agnostic features. }


Interestingly, the availability of multiple sources in the official challenge dataset make it perfect to tackle the problem under a DG setting. To this purpose, we extended RNA-Net to the Flow modality, obtaining remarkable results without accessing target data. In a second stage, we further adapted it to work with unlabelled target data under the standard Unsupervised Domain Adaptation (UDA) setting. Finally, our final submission was obtained by ensembling different model streams by means of DA-based consistency losses, namely Temporal Hard Norm Alignment (T-HNA) and Min-Entropy Consistency (MEC). 




\section{Our Approach}
In this section, we first describe the DG approach we used. 
Then, we illustrate its extension to unlabelled target data under the standard UDA framework. Finally, we repurpose existing DA-based losses to induce consistency between different architectures. 

\subsection{Domain Generalization}
The multi-source nature of the proposed challenge setting makes it perfect to deal with the domain shift using DG techniques. 
Thus, we first exploited a method which has been recently proposed to operate in this context, called Relative Norm Alignment (RNA) \cite{planamente2021crossdomain}.  
This methods consists in performing an \textit{audio-visual domain alignment} at feature-level by minimizing a cross-modal loss function ($\mathcal{L}_{RNA}$). The latter aims at minimizing the \textit{mean-feature-norm distance} between the audio and visual features norms among all the source domains, 
and it is defined as
\begin{equation}\label{formula:rna_1}
    \mathcal{L}_{RNA}=\left(\frac{\EX[h(X^v)]}{\EX[h(X^a)]} - 1\right)^2,
\end{equation}
where $h(x^m_i)=({\lVert{ \cdot }\rVert}_2 \circ f^m)(x^m_i)$ indicates the $L_2$-norm of the features $f^m$ of the $m$-th modality, $\EX[h(X^m)]=\frac{1}{N}\sum_{x^m_i \in \mathcal{X}^m}h(x^m_i)$ for the $m$-th modality and $N$ denotes the number of samples of the set $\mathcal{X}^m=\{x^m_1,...,x^m_N\}$.

Authors of \cite{planamente2021crossdomain} proved that the norm unbalance between different modalities might cause the model to be biased towards the source domain that generate features with greater norm and thus causing a wrong prediction.  
Indeed, by simultaneously solving the problem of classification and relative norm alignment on different domains, the network extracts a shared knowledge between the different sources,  
resulting in a domain-agnostic model. 

In our submission to the EPIC-Kitchen UDA challenge, we extended the RNA-Net framework to the optical flow modality, and we exploited the multiple sources available from the official training splits to show the effectiveness of RNA loss in a multi-source DG setting. 

    


\subsection{Domain Adaptation}
In this section, we describe the UDA techniques that are integrated in our approach.

\textbf{Relative Norm Alignment Network.} We followed the extension towards the UDA setting proposed in \cite{planamente2021crossdomain}, which is possible thanks to the unsupervised nature of RNA. 
In order to consider the contribution of both source and target data during training, we redefined $\mathcal{L}_{RNA}$ under the UDA setting as 

\begin{equation}\label{eq:loss_s_t}
    \mathcal{L}_{RNA}=\mathcal{L}^s_{RNA}+\mathcal{L}^t_{RNA} ,
\end{equation}
where $\mathcal{L}^s_{RNA}$ and $\mathcal{L}^t_{RNA}$ correspond to the RNA formulation in Equation \ref{formula:rna_1} illustrated above, when applied to source and target data respectively.

\textbf{Temporal Attentive Adversarial Adaptation Network (TA$^3$N).} Authors of \cite{videoda-chen2019temporal} proposed an UDA technique based on three components. The first one,  called \textit{Temporal Adversarial Adaptation Network (TA$^2$N)}, consists in an extension of DANN \cite{grl-pmlr-v37-ganin15}, 
aiming to align the temporal features on a multi-scale Temporal Relation Module (TRM) \cite{zhou2018temporal} through a gradient reversal layer (GRL).
The second component is based on a domain attention mechanism which guides the temporal alignment towards features where the domain discrepancy is larger.
Finally, the third component uses a minimum entropy regularization (attentive entropy) to
refine the classifier adaptation.


\setlength\heavyrulewidth{0.31ex}

\begin{table*}[t]
\centering
\begin{adjustbox}{width=0.85\linewidth, margin=0ex 1ex 0ex 0ex}
\begin{tabular}{l|c|cccccc}
\toprule\noalign{\smallskip}
\multicolumn{8}{c}{\normalsize\textsc{Unsupervised Domain Adaptation Leaderboard}} \\
\noalign{\smallskip}
\cline{1-8}
\noalign{\smallskip}
  & Rank & \multicolumn{1}{c}{Verb Top-1} & Noun Top-1 & Action Top-1& Verb Top-5 & Noun Top-5 & Action Top-5\\ 
 \noalign{\smallskip} \hline
 chengyi        &1          & 53.16      & 34.86      & \textbf{25.00}        & 80.74      & 59.30      & 40.75        \\ \hline
M3EM             &2        & 53.29      & \textbf{35.64 }     & 24.76        & 81.64      & 59.89      & 40.73        \\ \hline
\rowcolor[HTML]{AFDAAF} 
 plnet        & 3            &\textbf{ 55.22}      & 34.83      & 24.71        & \textbf{81.93 }     & \textbf{60.48}      & \textbf{41.41}        \\ \hline
 EPIC\_TA3N \cite{damen2020rescaling}      & 6      & 46.91      & 27.69      & 18.95        & 72.70      & 50.72      & 30.53        \\ \hline
 EPIC\_TA3N\_SOURCE\_ONLY  \cite{damen2020rescaling} & 12 & 44.39      & 25.30      & 16.79        & 69.69      & 48.40      & 29.06   \\ 
\bottomrule
\end{tabular}
\end{adjustbox}
\caption{Leaderboard results of EPIC-Kitchens Unsupervised Domain Adaptation Challenge. The results obtained by the top-3 participants and the provided baseline methods are reported. \textbf{Bold:} highest result; {\color{Green}{\textbf{Green:}}} our final submission. }
\label{leaderboard}
\end{table*}

\setlength{\tabcolsep}{5pt}
\begin{table*}[t]
\begin{minipage}{0.44\linewidth}
\begin{adjustbox}{width=1\linewidth, margin=0ex 0ex 0ex 0ex}
\begin{tabular}{l|ccc|ccc}
\toprule\noalign{\smallskip}
\multicolumn{7}{c}{\normalsize\textsc{Ensemble UDA losses}} \\
\noalign{\smallskip}
\cline{1-7}
\noalign{\smallskip}
   & \multicolumn{3}{c|}{Top-1} & \multicolumn{3}{c}{{Top-5}}\\ 
 \noalign{\smallskip}  \hline
 \noalign{\smallskip} 
  & \multicolumn{1}{c}{Verb} & Noun & Action & Verb  & Noun & Action \\  \hline \noalign{\smallskip}

{Ensemble}                       & 52.83      & 30.82      & 21.96        & \textbf{81.04}      & 52.67      & 46.66        \\ \hline \noalign{\smallskip}

Ensemble+T-HNA                 & 53.84      & 32.54      & 22.65        & 80.63      & 54.86      & 48.03        \\ \hline \noalign{\smallskip}

Ensemble+T-HNA+MEC & \textbf{54.02}      & \textbf{33.53 }     & \textbf{23.58 }       & 81.00      & \textbf{55.03}      & \textbf{48.27 }       \\ 
\bottomrule
\end{tabular}
\end{adjustbox}
\end{minipage}
\hspace{0.05\textwidth}
\centering
\begin{minipage}{0.35\linewidth}
\begin{adjustbox}{width=1\columnwidth, margin=0ex 0ex 0ex 0ex}
\begin{tabular}{l|ccc}
\toprule\noalign{\smallskip}
\multicolumn{4}{c}{\normalsize\textsc{Domain Generalization}} \\
\noalign{\smallskip}
\cline{1-4}
\noalign{\smallskip}
  & \multicolumn{1}{c}{Target} & Verb Top-1 & Verb Top-5 \\ 
 \noalign{\smallskip} \hline \noalign{\smallskip}

Source Only & \xmark               & 44.39 & 69.69 \\ \hline \noalign{\smallskip}

EPIC\_TA3N  \cite{damen2020rescaling}      & \cmark               & 46.91 & 72.70 \\ \hline \noalign{\smallskip}

RNA-Net \cite{planamente2021crossdomain}        & \xmark              & \underline{47.96} & \underline{79.54} \\ \hline
\noalign{\smallskip}
EPIC\_TA3N+RNA-Net    & \cmark               & \textbf{50.40} & \textbf{80.47} \\ 
\bottomrule
\end{tabular}
\end{adjustbox}
\caption{\textbf{Left.} Results on the EPIC-Kitchen validation set with different ensembling UDA losses. \textbf{Right.} Results on EPIC-Kitchen test set under the DG setting. \textbf{Bold} highest result. }
\label{multi-stream}
\end{minipage}
\end{table*}

\subsection{Ensemble UDA losses}

For our final submission, different models are used in order to exploit the potentiality of popular video architectures. 
Training individually each backbone with standard UDA protocols results in an adapted feature representation which varies from stream to stream. Our intuition is that this aspect could impact negatively the training process and the performance on target data. 
In fact, since the domain adaption process acts on each architecture independently, different prediction logits are obtained on target data. When combining them, this could cause a mismatch between the final scores, increasing the level of uncertainty of the model.
Thus, we impose a consistency constraint between feature representations from different models, by repurposing existing UDA loss functions to operate between multiple streams. Those are:

\textbf{Temporal Hard Norm Alignment (T-HNA).} It re-balances the contribution of each model during training by extending HNA \cite{planamente2021crossdomain} to align the norms of features coming from the different streams towards the same value $R$. This is applied on features extracted from multiple scales of each TRN module. The resulting $\mathcal{L}_\textit{T-HNA}$ is defined as
    
    \begin{equation}\label{eq:hna}
    \mathcal{L}_\textit{T-HNA}=\sum_b\left(\EX[h_t(X^b)] - R\right)^2,
\end{equation}
where $h_t$ denotes the $L_2$-norm of features extracted from the $t$-th multi-scale level of the $b$-th backbone network. 
    
\textbf{Min Entropy Consensus (MEC loss).} We extended the loss proposed in \cite{roy2019unsupervised} to encourage coherent predictions between different models. The resulting loss is defined as:
    \begin{equation}\label{eq:hna}
    \mathcal{L}_{MEC} = - \frac{1}{m}\sum_{i=1}^{m} \frac{1}{b} \max_{y \in \mathcal{Y} }{ \sum_{b} \mathrm{log} \textit{p}_b(y | x_{i}^{t})}
    \end{equation}
where $m$ is the cardinality of the batch size of the target set, $y$ is the predicted class, and $\mathrm{log} \textit{p}_b(y | x_{i}^{t})$ is the prediction probability of the $b$-th backbone network. The intuitive idea behind the proposed approach is to encourage different backbones to have a similar predictions.

\section{Framework}
In this section, we describe the architectures of the feature extractors used to produce suitable multi-modal video embeddings, and the fusion stategies adopted to combine them. We complete this section with the description of the hyper-parameters used for the training.

\subsection{Architecture}
\setlength\heavyrulewidth{0.31ex}

\textbf{Backbone.} 
For our submission, we adopted different network configurations.
In the first one, corresponding to the RNA-Net framework in \cite{planamente2021crossdomain}, 
we used the Inflated 3D ConvNet (I3D), pre-trained on Kinetics \cite{carreira2017quo}, for RGB and Flow streams, and a BN-Inception model \cite{ioffe2015batch} pre-trained on ImageNet \cite{imageNet} for the auditory information. 
Each feature extractor produces a 1024-dimensional representation which is fed to an action classifier. 
In the second configuration, we used BNInception for all the three streams, using pre-extracted features from a TBN \cite{munro2020multi} model trained on EPIC-Kitchens-55. In the last configurations, we used standard ResNet50 \cite{he2016deep} 
for all the streams using TSN \cite{wang2016temporal} and TSM~\cite{lin2019tsm} models pre-trained on Epic-Kitchen55\footnote{\url{https://github.com/epic-kitchens/epic-kitchens-55-action-models}}. 

\textbf{Multi-modal fusion strategies.}
In all the above mentioned configurations, each modality is processed by its own backbone, and the corresponding extracted representations are then fused following different strategies.
For RNA-Net, we followed a standard late fusion strategy, consisting in averaging the final score predictions obtained from two different fully-connected layers (verb, noun) from each modality. In the other configurations, we adopted the mid-fusion strategy proposed in \cite{Kazakos_2019_ICCV}, to generate a common frame-embedding among the modalities and used a Temporal Relation Module (TRM) \cite{zhou2018temporal} to aggregate features from different frames 
before feeding the final embeddings to the verb and noun classifiers.

\setlength{\tabcolsep}{10pt}

\begin{table}[t]
\centering
\begin{adjustbox}{width=\columnwidth, margin=0ex 1ex 0ex 0ex}
\begin{tabular}{|c|c|c|c|c|c|}
\hline
$\lambda_{RNA}$ & $\lambda_{HNA}$ & $R$ & $\lambda_{MEC}$ & $\gamma$ & $\beta$         \\ \hline
1             & 0.0006        & 40         & 0.01                     & 0.003    & 0.75, 0.75, 0.5 \\ \hline
\end{tabular}
\end{adjustbox}
\caption{UDA losses hyper-parameters used during training.} 
\label{params}
\end{table}

\subsection{Implementation Details}
We trained I3D and BNInception models with SGD optimizer, with an initial learning rate of 0.001, dropout 0.7, and using a batch size of 128, following \cite{planamente2021crossdomain}. Instead, when using pre-extracted features from ResNet50 or BNInception, we trained the TRM modules on top of them for 100 epochs with an initial learning rate of 0.03, decayed after epochs 30 and 60 by a factor of 0.1. We used a batch size of 128 with SGD optimizer. In Table \ref{params} we report the other hyper-parameter used. Specifically, we indicate with $\lambda_{RNA}$, $\lambda_{T-HNA}$ and $\lambda_{MEC}$ the weights of RNA, T-HNA and MEC losses respectively, and with $R$ the values of the radius of T-HNA (see Equation \ref{eq:hna}). In addition, we report the values used in TA$^3$N to weight the attentive entropy loss ($\gamma$) and the domain losses at different levels ($\beta$). 

\section{Results and Discussion}


In Table \ref{leaderboard} we report our best performing model on the target test, achieving the \textbf{1st} position on ‘verb', \textbf{3rd} on ‘noun' and ‘action', and \textbf{1st} position on Top-5 accuracy on all categories. In Table \ref{multi-stream} (left) we show an ablation on the contribution of the proposed ensemble UDA losses, T-HNA and MEC respectively, on the official validation set. As it can be seen, they improve Top-1 accuracy on all categories by up to $2\%$, proving the effectiveness of imposing a consistency between features from different streams. 

\textit{How well do DG approaches perform? } We show in Table \ref{multi-stream} (right) the results obtained under the multi-source DG setting, when target data are not available during training. Noticeably, RNA outperforms the baseline Source Only by up to $3\%$ on Top-1 and $10\%$ on Top-5, remarking the importance of using ad-hoc alignment techniques to deal with multiple sources in order to effectively extract a domain-agnostic model. Moreover, it outperforms the very recent UDA technique TA$^3$N without accessing to target data. 
Interestingly, when combined with EPIC\_TA3N, it further improves performance, proving the complementarity of RNA to other existing UDA approaches.

{\small
\bibliographystyle{ieee_fullname}
\bibliography{egbib}
}

\end{document}